\documentclass[sigconf,nonacm]{acmart}
\usepackage{amsmath}
\usepackage{booktabs}
\usepackage{graphicx}
\usepackage{tabularx}
\usepackage{array}
\usepackage{placeins}
\newcolumntype{Y}{>{\raggedright\arraybackslash}X}
\title{Account Consistency from Gameplay Traces: Same-Player Verification in Counter-Strike 2}
\author{Xuchen Zhang}
\affiliation{%
  \institution{Independent Researcher}
  \country{China}
}
\email{tigerlovezj@outlook.com}

\begin{document}
\begin{abstract}
In competitive first-person shooter (FPS) games such as Counter-Strike 2 (CS2), account-integrity review often asks whether an account's recent behavior remains consistent with its historical operator. This consistency question arises in cases such as temporary substitution, rank boosting, and high-skill players using lower-ranked accounts, where manual review requires comparing a current match against multiple historical matches. We formulate this review task as same-player verification: we encode the behavioral trajectory of a single player in a match replay (demo) as a demo-player behavioral fingerprint, and train a model to judge whether two behavioral observations come from the same real player. Using CS2-specific domain knowledge, the fingerprints cover crosshair control, movement-stop-fire coordination, economy/buy, combat/engagement, and temporal rhythm. We construct strict six-fold evaluations on the Perfect dataset ($\mathcal{D}_{\mathrm{PER}}$; 3,570 demos and 35,700 demo-player observations) and the Professional dataset ($\mathcal{D}_{\mathrm{PRO}}$; 539 demos and 5,390 demo-player observations). The final pairwise model reaches ROC AUCs of 0.926 and 0.956, respectively. Feature analysis shows that the strongest identity signals come from aiming/crosshair and other low-level mechanical behaviors, indicating that stable mechanics are more informative for this verification task than single-match performance outcomes. On fixed eligible query cohorts, aggregating pairwise evidence between a current demo and multiple historical demos raises account-history AUC on Perfect from 0.923 at $K=1$ to 0.982 at $K=10$, and on Professional from 0.914 at $K=1$ to 0.975 at $K=5$. These results show that CS2 demo behavior can support supervised same-player verification and account-level identity-consistency modeling through multi-demo history aggregation.
\end{abstract}
\maketitle
\hypersetup{%
  pdfauthor={Xuchen Zhang},
  pdftitle={Account Consistency from Gameplay Traces: Same-Player Verification in Counter-Strike 2},
  pdfsubject={Preprint},
  pdfkeywords={same-player verification, account consistency, behavioral biometrics, game telemetry, Counter-Strike 2}
}

\section{Introduction}
\subsection{Problem Background}
Counter-Strike 2 is one of the most active competitive FPS games \cite{valve-cs2-steam,steamdb-cs2-charts}.

Beyond anti-cheat, operator consistency within an account history is a distinct fairness concern in competitive FPS platforms. Account sharing, rank boosting, temporary player substitution, and high-skill players using another person's lower-ranked account can make current operator behavior inconsistent with the account's historical behavior, undermining matchmaking fairness, player trust, and tournament credibility. To handle such cases, platforms and tournament organizers also need to judge whether the operator's behavior in the current match is still consistent with the account's past behavioral patterns.

Existing platform mechanisms address admission, reporting, and case review, while account-history consistency requires a different comparison: current behavior against multiple historical matches. Identity verification suits account admission, tournament registration, or high-risk checkpoints, but is hard to trigger frequently across everyday matches; player reports are low-cost but noisy; and manual demo review can inspect single-match segments but cannot systematically compare the current match against multiple historical matches in shooting habits, mechanical habits, economy decisions, and round rhythm. As a result, existing workflows struggle to turn account-history consistency review into a routine workflow.

CS2 demos provide structured telemetry for modeling fine-grained player operations and decisions across rounds. Viewed more generally, this is a longitudinal user-activity modeling problem over structured platform telemetry, with open-set consistency verification as its target. This paper uses these structured behavioral trajectories to build demo-player behavioral fingerprints and judge the consistency of player behavior across matches.

\subsection{Task and Approach Overview}
We model account-history consistency review as open-set same-player verification. Here, open-set means that test-time players need not appear in training; the model does not identify who the current operator is, but learns a reusable comparison function that judges whether two segments of demo-player behavior come from the same real player. In account-history review, this function evaluates behavioral consistency between the current match and historical matches and forms account-level identity-consistency evidence.

We model operation and decision patterns at a lower level than K/D (kill/death ratio), ADR (average damage per round), headshot rate, or rank, such as crosshair micro-adjustment, firing rhythm, movement-stop-fire coordination, and buying preference. Buying preference can be adjusted intentionally, while low-level operations and action-timing coordination such as crosshair control, firing rhythm, and movement-stop-fire coordination are less directly controllable and may be harder to imitate consistently across rounds.

The per-demo-player fingerprint has two parts: aggregate behavioral fingerprint features constructed from game understanding, characterizing stable behaviors such as crosshair, movement, economy, combat, and timing; and Transformer-derived sequence embeddings used as complementary behavioral representations for action order, state switching, and low-level mechanics from round events and combat windows. The pairwise model uses both representations, and account-history review aggregates current-vs-history scores into an account-level consistency signal.

\subsection{Contributions}
This paper makes the following contributions:

\begin{itemize}
\item \textbf{A supervised formulation for FPS account-history consistency.} We formulate account-history consistency review as a same-player verification task that a model can learn.
\item \textbf{Behavioral findings from CS2 fingerprints.} We find that player identity signals come mainly from aiming/crosshair and other low-level mechanical behaviors; sequence modeling captures complementary action-order information, with dataset-dependent gains.
\item \textbf{Verification and aggregation under person-disjoint evaluation.} We evaluate the formulation across person-disjoint splits and account-history aggregation settings, extending single-pair verification to multi-demo account-history consistency modeling.
\end{itemize}

\section{Related Work}
\subsection{Account Identity Verification and Fairness Mechanisms on Competitive Platforms}
Competitive platforms have already incorporated account integrity into fairness governance. Both operator inconsistency behind an account and one person using multiple accounts to bypass platform rules can undermine matchmaking fairness and tournament credibility. Existing esports research discusses boosting---``finding a stronger player to play on an account to improve its rank or results''---and the fairness risk from inconsistency between an account's displayed skill and the real operator's ability \cite{conroy-boosting-2021}; broader research on online-game cheating shows that platform governance often needs to combine multiple signals such as accounts, behavior, and social relations \cite{blackburn-toit-2014-cheating-social}.

Mainstream platforms combine anti-cheat, identity verification, reporting, and manual review to manage account-integrity risks \cite{faceit-banning-policy,faceit-verification-process}. These mechanisms support admission control, user reports, and case review, but they do not directly provide a systematic comparison between current behavior and multiple historical matches. This paper formalizes that longitudinal current-vs-history comparison as CS2 demo-based same-player verification.

\subsection{Counter-Strike and FPS Player Behavioral Identity Modeling}
The closest game-domain work studies Counter-Strike / CS2 player identity and fair-play. Existing work uses behavioral features to identify known professional players or distinguish known player pairs \cite{zimmer-hicss-2025-csgo-player-identity}, and related CoG work studies in-game behavioral biometrics for fair play \cite{zimmer-cog-2025-fair-play-identity}. These studies show that CS/CS2 demo telemetry contains identity-related signals, especially in aiming, shooting, movement state, and game context.

Our setting is different: platform review often asks whether a current demo remains consistent with an account's historical demos when the current operator may be unknown or unseen during training. We therefore use manually confirmed same-player histories, focus on stable CS2 habits such as crosshair control, firing rhythm, movement-stop-fire coordination, state switching, buying rhythm, and risk preference, and evaluate with unseen-player splits plus current-vs-history aggregation. This moves the target from fixed-identity recognition to account-history consistency.

Another related direction studies FPS fair-play risk and skill/rating discrepancy. GUARD uses mouse/keyboard dynamics, in-game actions, and expert knowledge to infer a player's skill group, then compares it with the account rating to identify smurfing / rank-boosting risk \cite{guard-eswa-2026}. FPS security research also detects passive aimbots through inconsistency between shooting performance and broader skillfulness \cite{liu-dsn-2017-passive-cheats}. These methods can flag cases where operator ability clearly mismatches account rank, or where shooting performance is inconsistent with broader skillfulness. Account borrowing, temporary substitution, or short-term boosting can also happen without an obvious ability jump. We therefore evaluate behavioral consistency between the current demo and account-history demos, directly comparing whether behavior before and after still follows the same player's operational habits.

\subsection{Broader Game Behavior and Behavioral Biometrics}
Beyond FPS games, replay and telemetry have also been used for player identity and style modeling. Dota 2 work takes whether two matches were completed by the same player as the target and models it with mouse, game statistics, and strategy information \cite{yuen-dota2-player-identification-2020}. RTS (real-time strategy) replay identification shows that build order, unit control, resource management, and operation rhythm can identify player style or identity \cite{liu-rts-replay-2013}.

Behavioral biometrics research shows that identity signals can come from how a person acts---mouse trajectories, touchscreen habits, interaction rhythms, and motion patterns \cite{behavioral-biometrics-infofusion-2021,smartphone-behavioral-biometrics-2017}; VR head/hand motion, mouse dynamics in simple cognitive games, and active-user identification under shared accounts report similar findings \cite{nair-usenix-2023-vr-identification,gametrics-acsac-2016,time-aware-user-identification-icdm-2016}. These studies provide background evidence that behavioral telemetry can carry identity signal, but their settings differ from natural CS2 account-history review. This paper focuses on the CS2 account-history consistency setting, where identity evidence comes from FPS-specific motor, timing, and tactical behavior.

\subsection{Positioning of This Paper}
The above lines of work are complementary to this paper. Platform mechanisms provide admission control, reporting, and manual review entry points; Counter-Strike / FPS behavior studies show that in-game behavior contains identity signal; broader behavioral biometrics shows that identity evidence can emerge from how people act. We instantiate these ideas in CS2 account-history consistency review: using manually confirmed account histories to construct pairwise labels, learning same-player verification for unseen players, and identifying crosshair control, movement-stop-fire coordination, and combat micro-operations as the behavioral signals that most strongly support this judgment.

\section{Method}
\subsection{Problem Definition}
We call a parsable behavioral record left by one player in one CS2 match demo a \textbf{demo-player observation}; a standard CS2 match demo usually contains 10 players and therefore produces 10 demo-player observations. A demo records structured replay / telemetry from the game engine, including positions, view angles, weapons, events, and round states.

We decompose the account-history consistency problem into two levels.

\textbf{Pairwise verification primitive.} The input is a pair of demo-player observations, where each observation represents the behavioral trajectory left by one player in one CS2 match demo. The verification model outputs an identity-consistency score, where higher scores indicate stronger evidence that the two observations come from the same real player.

\textbf{Account-history aggregation.} Given a current observation under review and \(K\) historical observations from the account, we compute \(K\) pairwise scores and aggregate them into an account-level consistency signal.

Formally, let \(x_i^{\mathrm{beh}}\) denote the game-understanding-based aggregate behavioral fingerprint of the \(i\)-th demo-player observation, whose construction is described in Section 3.2; let \(x_i^{\mathrm{seq}}\) denote the combat-window sequence representation produced by the sequence encoder in Section 3.3. We take their concatenation

\begin{equation}
x_i=[x_i^{\mathrm{beh}}, x_i^{\mathrm{seq}}]
\label{eq:fingerprint}
\end{equation}

as the complete behavioral fingerprint. Let \(p_i\) denote the real-player label to which this observation belongs. Player labels are used only to construct training and evaluation samples; at test time, the model does not need to identify who \(p_i\) is. For any pair of observations, we define the pair label:

\begin{equation}
y_{ij} = \mathbf{1}[p_i = p_j],
\label{eq:pair-label}
\end{equation}

where \(y_{ij}=1\) denotes a same-player pair and \(y_{ij}=0\) denotes a different-player pair. The model receives the endpoint fingerprints, the explicit comparison features \(\operatorname{compare}(x_i,x_j)\) defined in Section 3.4, and the pair-level context \(c_{ij}\), and outputs a consistency score:

\begin{equation}
\begin{aligned}
s_{ij}=g_\theta(&[x_i,x_j,\operatorname{compare}(x_i,x_j),c_{ij}]).
\end{aligned}
\label{eq:pair-score}
\end{equation}

where \(g_\theta\) is the pairwise comparator to be learned and \(\theta\) denotes model parameters; \(s_{ij}\) is the identity-consistency score. During training, \(y_{ij}\) supervises \(g_\theta\). A higher score indicates stronger same-player consistency, and different-player retrieval uses the low-score side.

In this paper, \(c_{ij}\) only encodes map relation, such as same-map versus cross-map; it does not include demo IDs, match IDs, teammate/opponent identities, or other shared match identifiers.

\subsection{Game-Understanding-Based Behavioral Fingerprint Features}
This section describes the construction of \(x_i^{\mathrm{beh}}\): from each demo-player observation we extract behavioral fingerprints based on CS2 game understanding to summarize the player's operation and decision habits in one match.

From the demo record we recover the player's within-round position, view angle, movement state, weapon state, firing, damage, utility, and buy events, and express these behaviors at demo-player granularity as \(x_i^{\mathrm{beh}}\): a summary of behavioral frequency, time intervals, distribution shapes, and conditional relations that characterizes how a player moves, aims, fires, switches states, buys equipment, and uses utility across maps and round phases. In implementation, we extract 245 per-demo-player behavioral fingerprint features and organize them into eight sub-representations:

\begin{equation}
\begin{aligned}
x_i^{\mathrm{beh}}=[&
x_i^{\mathrm{beh}\text{-}\mathrm{aim}},x_i^{\mathrm{beh}\text{-}\mathrm{mech}},x_i^{\mathrm{beh}\text{-}\mathrm{combat}},x_i^{\mathrm{beh}\text{-}\mathrm{move}},\\
&x_i^{\mathrm{beh}\text{-}\mathrm{util}},x_i^{\mathrm{beh}\text{-}\mathrm{time}},x_i^{\mathrm{beh}\text{-}\mathrm{econ}},x_i^{\mathrm{beh}\text{-}\mathrm{ctx}}].
\end{aligned}
\label{eq:behavior-families}
\end{equation}

\begin{table*}[t]
\centering
\footnotesize
\setlength{\tabcolsep}{3pt}
\renewcommand{\arraystretch}{1.12}
\caption{Game-understanding-based behavioral fingerprint feature families and quantification examples.}
\label{tab:1}
\begin{tabularx}{\textwidth}{@{}>{\raggedright\arraybackslash}p{0.09\textwidth}>{\raggedright\arraybackslash}p{0.075\textwidth}>{\raggedright\arraybackslash}p{0.13\textwidth}>{\raggedright\arraybackslash}p{0.29\textwidth}>{\raggedright\arraybackslash}X@{}}
\toprule
behavioral layer & symbol & feature family & behavior captured & concrete quantification example \\
\midrule
low-level & \(x_i^{\mathrm{beh}\text{-}\mathrm{aim}}\) & aiming/crosshair & view control, correction, recoil, aiming stability & left-right crosshair correction switches around firing \\
low-level & \(x_i^{\mathrm{beh}\text{-}\mathrm{mech}}\) & mechanics/state & counter-strafe, walk/crouch/scope, state switch & speed drop in the 250ms before each shot \\
low-level & \(x_i^{\mathrm{beh}\text{-}\mathrm{combat}}\) & combat/engagement & shooting discipline, reload rhythm, fire output & reloads per 100 weapon fires \\
rhythm-space & \(x_i^{\mathrm{beh}\text{-}\mathrm{move}}\) & movement/positioning & opening route, position, map-space preference & concentration of frequent opening positions \\
rhythm-space & \(x_i^{\mathrm{beh}\text{-}\mathrm{util}}\) & utility usage & utility timing, type choice, follow-up & damage within 5 seconds after utility release \\
rhythm-space & \(x_i^{\mathrm{beh}\text{-}\mathrm{time}}\) & timing/rhythm & first contact, firing interval, push/wait rhythm & seconds from round start to first weapon fire \\
tactical & \(x_i^{\mathrm{beh}\text{-}\mathrm{econ}}\) & economy/buy & buy choice and risk under economic pressure & force-buy rate under insufficient economy \\
tactical & \(x_i^{\mathrm{beh}\text{-}\mathrm{ctx}}\) & context & early risk, man advantage/disadvantage, role context & deaths in the first 20 seconds of a round \\
\bottomrule
\end{tabularx}
\end{table*}

Taking mechanics/state as an example, this feature type records not ``how many duels were won'' but how the player's body state changes before firing. In CS2, stable shooting usually requires counter-strafing; some players fire only after completely stopping, while others fire early while moving. Aggregating such micro-transition habits over the full demo-player observation expresses a player's long-term mechanical style.

\subsection{Sequential Behavioral Representation}
The behavioral fingerprint features in Section 3.2 efficiently summarize behavioral distributions that repeatedly appear in one match. However, similar firing counts, movement speeds, or utility counts can come from completely different round developments. For example, around the same combat, counter-strafing first and then micro-adjusting and firing, versus firing early while moving and then counter-strafing and micro-adjusting, reflect different movement-stop-fire coordination but may look almost identical in the features above; likewise, after throwing a utility item, immediately pushing, waiting for a teammate to trade, or only delaying tempo represent different utility-combat coordination.

We therefore encode action sequences, but not all ticks of the entire demo. A CS2 demo contains many rounds, and large portions of a demo may contain weak identity signal; player identity is more concentrated in short operations around firing, taking damage, dealing damage, kills, and the moments before death. We thus extract local windows centered on combat events from each demo-player observation: around firing, dealing/taking damage, kills, and death events, we crop fixed-length tick sequences, with death-related evidence concentrated before the event. Each window is a \(32\times16\) continuous numerical tensor whose channels are relative time, yaw/pitch velocities, yaw/pitch deltas, speed, horizontal velocity in two axes, displacement, tick interval, health, duck amount, walking and scoped indicators, shots fired, and a center-band indicator. The final sequence encoder retains valid tokens with \(\mathrm{shots\_fired}>0\); if an otherwise available observation has no selected firing token, it falls back to the base valid-token mask.

Let \(w_{im}\) denote the \(m\)-th combat window in observation \(i\). The sequence encoder \(E_\psi\) maps each window to a window embedding \(h_{im}\), and an aggregation function \(\operatorname{aggregate}_{\omega}\) summarizes them into a demo-player-level sequence embedding:

\begin{equation}
h_{im}=E_\psi(w_{im}),\qquad
x_i^{\mathrm{seq}}=\operatorname{aggregate}_{\omega}\left(\{h_{im}\}_{m=1}^{M_i}\right).
\label{eq:sequence-embedding}
\end{equation}

where \(M_i\) is the number of available combat windows. We use a two-layer, four-head Transformer with hidden size 96 and dropout 0.15. Learned attention pooling, masked mean, and masked standard deviation are concatenated and projected to a 192-dimensional observation embedding. The encoder minimizes identity cross-entropy plus 0.35 supervised contrastive loss (temperature 0.12), using at most 24 windows during training and 64 during evaluation.

\subsection{Pairwise Comparison Representation}
After obtaining \(x_i^{\mathrm{beh}}\) and \(x_i^{\mathrm{seq}}\), we concatenate them into the complete fingerprint \(x_i=[x_i^{\mathrm{beh}}, x_i^{\mathrm{seq}}]\). If two raw fingerprints are simply concatenated and fed to the model, the model must infer both feature differences and endpoint levels from limited samples. We therefore add a set of symmetric comparison features that explicitly express relations such as absolute and relative differences between two demo-players on the same behavioral dimensions.

Specifically, given two raw fingerprints \(x_i\) and \(x_j\), we construct explicit comparison features \(\operatorname{compare}(x_i,x_j)\) by concatenating behavior and sequence comparison blocks. For each scalar behavioral feature \(k\), the behavioral comparison block contains:

\begin{equation}
\begin{aligned}
\operatorname{compare}_k(x_i^{\mathrm{beh}},x_j^{\mathrm{beh}})=
\bigl[&|x_{ik}^{\mathrm{beh}}-x_{jk}^{\mathrm{beh}}|,\;
\frac{|x_{ik}^{\mathrm{beh}}-x_{jk}^{\mathrm{beh}}|}{|x_{ik}^{\mathrm{beh}}|+|x_{jk}^{\mathrm{beh}}|+\epsilon},\\
&\frac{x_{ik}^{\mathrm{beh}}+x_{jk}^{\mathrm{beh}}}{2},\;
\mathbf{1}[x_{ik}^{\mathrm{beh}}=0 \land x_{jk}^{\mathrm{beh}}=0],\\
&\mathbf{1}[(x_{ik}^{\mathrm{beh}}=0) \oplus (x_{jk}^{\mathrm{beh}}=0)]\bigr].
\end{aligned}
\label{eq:behavior-compare}
\end{equation}

where \(\epsilon\) ensures numerical stability and \(\mathbf{1}[\cdot]\) is the indicator function (1 when the bracketed condition holds, 0 otherwise); \(\oplus\) denotes exclusive OR. For the sequence embeddings, the comparison block contains the elementwise absolute difference, elementwise product, Euclidean distance, and cosine similarity:

\[
\begin{aligned}
\operatorname{compare}^{\mathrm{seq}}(x_i,x_j)=\bigl[&
|x_i^{\mathrm{seq}}-x_j^{\mathrm{seq}}|,\\
&x_i^{\mathrm{seq}}\odot x_j^{\mathrm{seq}},\;
\|x_i^{\mathrm{seq}}-x_j^{\mathrm{seq}}\|_2,\\
&\cos(x_i^{\mathrm{seq}},x_j^{\mathrm{seq}})\bigr].
\end{aligned}
\]

\subsection{Account-History Aggregation}
Actual account-consistency checks usually compare not just two demos, but one current demo under review against multiple historical demos of the account. Based on the \(s_{ij}\) defined in Section 3.1, we compare the current observation \(x_q\) with each of the account's \(K\) historical observations \(H=\{x_{h_1},\ldots,x_{h_K}\}\), obtaining \(K\) scores that describe the consistency between current behavior and account-history behavior:

\begin{equation}
\begin{aligned}
s_{q,h_k}=g_\theta(&[x_q,x_{h_k},\operatorname{compare}(x_q,x_{h_k}),c_{q,h_k}]),\\
&k=1,\ldots,K.
\end{aligned}
\label{eq:history-pair-score}
\end{equation}

Multi-demo aggregation converts \(K\) pairwise scores into an account-level consistency score. The most direct approach aggregates raw scores, such as the raw-score mean:

\begin{equation}
S_{\text{mean}}(q,H)=\frac{1}{K}\sum_{k=1}^{K}s_{q,h_k}.
\label{eq:raw-score-mean}
\end{equation}

We also evaluate an empirical LLR-style score transformation as an interpretable evidence scale for adding multiple pairwise scores. Based on the same / different score distributions on the training side, each pair score \(s\) is mapped to an evidence value \(\ell(s)\):

\begin{equation}
\ell(s)=\log\frac{\Pr(s\in b(s)\mid y=1)+\alpha}{\Pr(s\in b(s)\mid y=0)+\alpha},
\label{eq:llr-mapping}
\end{equation}

where \(b(s)\) is one of 20 fixed equal-width bins on \([0,1]\) and \(\alpha=1\). The bin distributions are estimated only from frozen fold-local validation predictions. Intuitively, \(\ell(s)\) is positive when a score bin is more common among same-player pairs and negative when it is more common among different-player pairs. We then average the evidence over the \(K\) historical observations:

\begin{equation}
S_{\mathrm{LLR}}(q,H)=\frac{1}{K}\sum_{k=1}^{K}\ell(s_{q,h_k}).
\label{eq:llr-mean}
\end{equation}

\subsection{Overall Workflow}
Fig. 1 shows the main data flow at inference time: each demo-player observation is first encoded as a fingerprint, two fingerprints are compared to obtain \(s_{ij}\), and multiple current-vs-history scores are then aggregated into an account-level consistency score. Training mainly supervises the pairwise comparator \(g_\theta\).

\begin{figure*}[t]
\centering
\includegraphics[width=\textwidth]{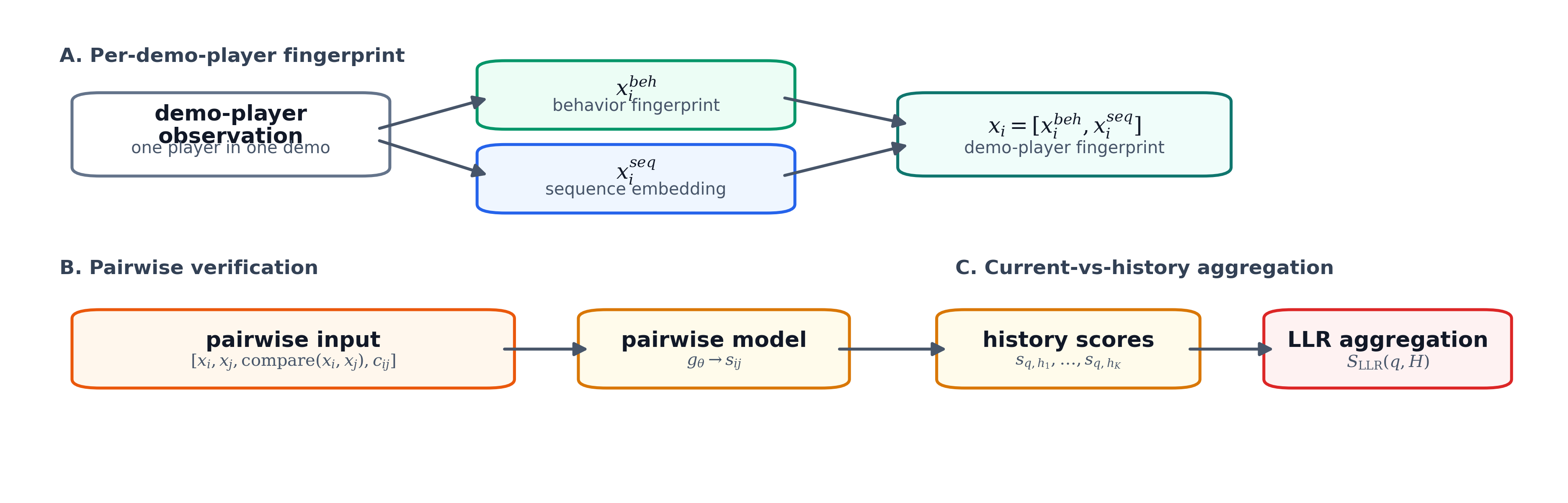}
\Description{Method overview showing behavioral fingerprint extraction, sequence encoding, pairwise comparison, and multi-demo history aggregation.}
\caption{Method overview: per-demo-player fingerprinting, pairwise scoring, and current-vs-history aggregation.}
\label{fig:1}
\end{figure*}

\section{Experiments}
\subsection{Dataset and Evaluation Setup}
\textbf{Perfect dataset and manual confirmation.} The Perfect dataset, $\mathcal{D}_{\mathrm{PER}}$, uses active users from a university CS player guild on Perfect World Arena~\cite{perfect-world-esports} as the collection entry point. We downloaded their available CS2 match demos within a specified time window, forming 3,570 demos and 35,700 demo-player observations. The guild players cover 12 amateur competitive tiers from C to Diamond S and above. To obtain credible same-player positives, we contacted active users in the guild and asked them to confirm whether the account was used only by themselves within the window, whether multiple accounts existed, and whether those accounts were all operated by the same real player. Observations with clear account borrowing, non-self play, uncertainty, or account-sharing risk do not enter same-player positives.

\textbf{Positive construction.} The 4,495 manually confirmed demo-player observations come from 107 confirmed persons and are used to construct same-player positives. For each confirmed person with \(n_p\) available observations, we enumerate \(\binom{n_p}{2}\) same-player combinations; across the six test folds, 171,713 same-player pairs are formed.

\textbf{Negative sampling.} Different-player pairs are sampled between observations with different person ids; observations manually confirmed to belong to the same real player or the same account group are first assigned to the same person id so they are not sampled as negatives. To cover negatives of different difficulty, the initial candidate quota assigns 25\% to same-demo pairs, 25\% to same-rank pairs, and 50\% to other randomly sampled different-player pairs; if a stratum lacks sufficient legal capacity, its shortfall is deterministically reassigned to the random stratum. The 35,700 observations theoretically form \(\binom{35700}{2}=637.2\)M possible pairs, the vast majority different-player. We keep all 171,713 same-player pairs in the six test folds and sample different-player pairs at an overall 1:9 ratio, obtaining 1,717,130 pairwise samples.

\textbf{Professional dataset.} In addition to $\mathcal{D}_{\mathrm{PER}}$, we downloaded public professional match demos from HLTV match/demo pages~\cite{hltv-matches} and built the Professional dataset, $\mathcal{D}_{\mathrm{PRO}}$. This set contains 539 professional match demos and 5,390 demo-player observations, among which 1,330 target professional-player observations correspond to 130 professional players. Professional identities come from public tournament records, providing externally verifiable longitudinal player records and coverage of elite play.

\textbf{Positive and negative construction.} In $\mathcal{D}_{\mathrm{PRO}}$, the 1,330 target observations used for positive construction come from 130 professional players; for each player $p$ with $n_p$ available observations, we likewise enumerate $\binom{n_p}{2}$ same-player combinations, yielding 10,769 same-player pairs across the six test folds. The 5,390 observations theoretically form $\binom{5390}{2}=14.5$M possible pairs, the vast majority different-player. Different-player pairs use the same initial stratification targets and capacity-shortfall reassignment rule. We retain all 10,769 same-player pairs in the six test folds and sample different-player pairs at an overall 1:9 ratio, yielding 107,690 pairwise samples. Across both datasets, only confirmed/target observations generate same-player positives; remaining roster observations enter only as different-player candidates under the frozen SteamID/alias mapping. Unlinked SteamIDs are treated as distinct identity units, so undisclosed cross-account ownership can create false-negative labels. For both datasets, AP is reported on the sampled evaluation distribution; deployment thresholds should be recalibrated on platform-specific data.

\textbf{Dataset use.} We evaluate representation levels, model comparison, feature sensitivity, history aggregation, and cross-time and cross-map robustness on both $\mathcal{D}_{\mathrm{PER}}$ and $\mathcal{D}_{\mathrm{PRO}}$, and further examine cross-dataset training. We additionally collect 513 demos and 5,130 demo-player observations from an independent player guild on the 5E platform~\cite{five-e-platform}, forming the 5E dataset $\mathcal{D}_{5\mathrm{E}}$. We use $\mathcal{D}_{5\mathrm{E}}$ as a cross-platform external test set with weak account labels to evaluate zero-shot generalization of models trained on Perfect.

\textbf{Split protocol.} Both datasets first form six folds by the known real-player identity ledger and then assign each demo to one side. Under this ledger, natural-person, SteamID, alias, observation, demo, and content overlaps between training and test are zero; undisclosed cross-account ownership remains possible label noise. Pairs crossing the two sides are discarded. Training and validation pairs follow the same legal constraints and 1:9 same-to-different ratio as the test pairs. Within each outer fold, sequence encoders and pairwise scorers fit only training-side identities; validation selects checkpoints and iteration counts, and the test fold is used only for final scoring. Pair endpoints use ascending frozen endpoint index as the canonical order.

\textbf{Map factor.} CS2 competitive matches concentrate on a small active-duty competitive map pool, including Dust2 and Mirage. Maps affect default routes, combat distances, and utility combinations; still, the same player is likely to retain stable habits such as crosshair control, firing rhythm, movement-stop-fire coordination, and risk preference across maps. We therefore verify both same-map and cross-map pairs.

\textbf{Evaluation sets.} Table~\ref{tab:2} gives the evaluation views used in this paper. The E1--E5 representation ladder and headline final-model results use the sequence-common surfaces $\mathcal{T}_{\mathrm{PER},pair,seq}^{(1:6)}$ and $\mathcal{T}_{\mathrm{PRO},pair,seq}^{(1:6)}$, where all representation levels share identical pair IDs, labels, and order. The full pair surfaces $\mathcal{T}_{\mathrm{PER},pair}^{(1:6)}$ and $\mathcal{T}_{\mathrm{PRO},pair}^{(1:6)}$ are used for non-sequence analyses; $\mathcal{T}_{\mathrm{PER},pair}^{(1)}$ is used only as the Perfect grouped-feature analysis split, whereas the corresponding Professional analysis uses all six folds. $\mathcal{T}_{\mathrm{PER},hist}^{(1:6)}(K)$ and $\mathcal{T}_{\mathrm{PRO},hist}^{(1:6)}(K)$ test whether aggregating the current demo with $K$ historical demos forms more stable account-level evidence; $\mathcal{T}_{\mathrm{PER},time}^{(1:6)}$ and $\mathcal{T}_{\mathrm{PRO},time}^{(1:6)}$ measure degradation over longer time spans; and $\mathcal{T}_{\mathrm{PER},map}^{(1:6)}$ and $\mathcal{T}_{\mathrm{PRO},map}^{(1:6)}$ examine robustness in cross-map comparison.

\begin{table*}[t]
\centering
\scriptsize
\setlength{\tabcolsep}{2.5pt}
\renewcommand{\arraystretch}{1.08}
\caption{Evaluation sets and metrics used in the experiments.}
\label{tab:2}
\begin{tabularx}{\textwidth}{@{}>{\raggedright\arraybackslash}p{0.16\textwidth}>{\raggedright\arraybackslash}p{0.17\textwidth}>{\raggedright\arraybackslash}p{0.24\textwidth}>{\raggedright\arraybackslash}p{0.20\textwidth}X@{}}
\toprule
symbol & purpose & construction & size & metrics \\
\midrule
$\mathcal{T}_{\mathrm{PER},pair}^{(1:6)}$ & full Perfect pair surface & six person-first/demo-ownership folds; known-identity, SteamID, alias, observation, demo, and content overlap is zero & total $N=1{,}717{,}130$ = 171,713 same + 1,545,417 different & mean AUC, same-AP \\
$\mathcal{T}_{\mathrm{PRO},pair}^{(1:6)}$ & full Professional pair surface & the same strict six-fold protocol & total $N=107{,}690$ = 10,769 same + 96,921 different & mean AUC, same-AP \\
$\mathcal{T}_{\mathrm{PER},pair}^{(1)}$ & Perfect grouped feature sensitivity & first test fold of $\mathcal{T}_{\mathrm{PER},pair}^{(1:6)}$ & $N=286{,}170$ = 28,617 same + 257,553 different & sensitivity AUC \\
$\mathcal{T}_{\mathrm{PER},pair,seq}^{(1:6)}$ / $\mathcal{T}_{\mathrm{PRO},pair,seq}^{(1:6)}$ & E1--E5 ladder and headline final model & identical sequence-common pair IDs, labels, and order within each dataset & Perfect: 1,518,330 pairs; Professional: 107,690 pairs & E1--E5 mean AUC / same-AP \\
$\mathcal{T}_{\mathrm{PER},hist}^{(1:6)}(K)$ / $\mathcal{T}_{\mathrm{PRO},hist}^{(1:6)}(K)$ & strict-prior history aggregation & current observation with $K$ earlier history observations & Perfect: $K=1,3,5,10$; Professional: $K=1,3,5$ & account mean-LLR AUC \\
$\mathcal{T}_{\mathrm{PER},time}^{(1:6)}$ / $\mathcal{T}_{\mathrm{PRO},time}^{(1:6)}$ & cross-time comparison & same day, 1--7, 8--30, 31--90, and $>90$ days & available buckets in frozen six-fold predictions & AUC \\
$\mathcal{T}_{\mathrm{PER},map}^{(1:6)}$ / $\mathcal{T}_{\mathrm{PRO},map}^{(1:6)}$ & map robustness & split frozen predictions by map relation & Perfect: 462,043 same-map + 1,056,287 cross-map; Professional: 38,295 + 69,395 & AUC, same-AP \\
$\mathcal{T}_{5\mathrm{E}}$ & cross-platform weak-account test & exact SteamID as weak same-account label & $N=55{,}840$ = 5,584 same + 50,256 different & AUC \\
$\mathcal{T}_{5\mathrm{E}}^{\mathrm{disjoint}}$ & exact-SteamID-disjoint sensitivity & exclude one SteamID overlapping the source & $N=55{,}676$ = 5,583 same + 50,093 different & AUC \\
\bottomrule
\end{tabularx}
\end{table*}

In pairwise tables, AP (Average Precision) uses same-player as the positive class. On the 5E weak-label surface, AP analogously uses same-account as the positive class.

Reported means and deltas are calculated from unrounded fold scores.

\subsection{Behavioral Representations and Model Comparison}
\subsubsection{Representation Levels: Which Information Brings Gains}
Table~\ref{tab:3} uses the averages over $\mathcal{T}_{\mathrm{PER},pair,seq}^{(1:6)}$ and $\mathcal{T}_{\mathrm{PRO},pair,seq}^{(1:6)}$ as the main pairwise results for the two datasets. The main $\mathcal{D}_{\mathrm{PRO}}$ results use a sequence encoder trained and frozen on the larger $\mathcal{D}_{\mathrm{PER}}$, avoiding a sequence representation determined only by the smaller Professional training set. On each split, the same LightGBM pairwise model \cite{lightgbm-nips-2017} compares how much same-player consistency signal different input representations provide. E1 uses \(x^{\mathrm{out}}\), i.e., outcome/performance-only features, including result-based statistics such as K/D (kill/death ratio), damage, score (game scoreboard score), headshot rate, and first kill / first death. E5 corresponds to the full pairwise input in Eq.~\eqref{eq:pair-score} in Section 3.1.

\begin{table}[t]
\centering
\tiny
\setlength{\tabcolsep}{1.15pt}
\renewcommand{\arraystretch}{1.05}
\caption{Representation ladder on the two sequence-common evaluation surfaces.}
\label{tab:3}
\begin{tabular*}{\columnwidth}{@{\extracolsep{\fill}}ccccc@{}}
\toprule
data & exp. & input & AUC & same-AP \\
\midrule
$\mathcal{D}_{\mathrm{PER}}$ & E1 & outcome & 0.572 & 0.130 \\
$\mathcal{D}_{\mathrm{PER}}$ & E2 & behavior & 0.855 & 0.452 \\
$\mathcal{D}_{\mathrm{PER}}$ & E3 & sequence & 0.719 & 0.227 \\
$\mathcal{D}_{\mathrm{PER}}$ & E4 & behavior + sequence & 0.859 & 0.454 \\
$\mathcal{D}_{\mathrm{PER}}$ & E5 & + explicit compare & \textbf{0.926} & \textbf{0.703} \\
$\mathcal{D}_{\mathrm{PRO}}$ & E1 & outcome & 0.712 & 0.210 \\
$\mathcal{D}_{\mathrm{PRO}}$ & E2 & behavior & 0.874 & 0.464 \\
$\mathcal{D}_{\mathrm{PRO}}$ & E3 & sequence & 0.749 & 0.254 \\
$\mathcal{D}_{\mathrm{PRO}}$ & E4 & behavior + sequence & 0.874 & 0.480 \\
$\mathcal{D}_{\mathrm{PRO}}$ & E5 & + explicit compare & \textbf{0.956} & \textbf{0.775} \\
\bottomrule
\end{tabular*}
\end{table}

E1, using only result-based statistics, is a weak baseline, with AUCs of 0.572 and 0.712 on $\mathcal{D}_{\mathrm{PER}}$ and $\mathcal{D}_{\mathrm{PRO}}$, respectively. Using the CS2-understanding-based $x^{\mathrm{beh}}$ in E2 yields AUCs of 0.855 and 0.874 on $\mathcal{T}_{\mathrm{PER},pair,seq}^{(1:6)}$ and $\mathcal{T}_{\mathrm{PRO},pair,seq}^{(1:6)}$, respectively, showing that crosshair control, movement-stop-fire coordination, combat rhythm, economy/buy, and round timing contain strong player-identity signals. E3, using the sequence representation $x^{\mathrm{seq}}$ alone, reaches AUCs of 0.719 and 0.749 on the two surfaces, indicating that action order in combat windows carries identity information but does not alone cover behaviors that recur across rounds. The main $\mathcal{D}_{\mathrm{PRO}}$ configuration uses the encoder pretrained and frozen on $\mathcal{D}_{\mathrm{PER}}$; an alternative trained only on $\mathcal{D}_{\mathrm{PRO}}$ training data yields a nearly identical E3 AUC of 0.747 versus 0.749.

Compared with E2, E4 combines $x^{\mathrm{beh}}$ and $x^{\mathrm{seq}}$: on $\mathcal{D}_{\mathrm{PER}}$, AUC improves by 0.004 on average (+0.005/+0.002/+0.006/-0.011/+0.004/+0.018); on $\mathcal{D}_{\mathrm{PRO}}$, AUC changes by 0.000 on average (-0.017/+0.005/-0.005/-0.002/+0.007/+0.014). Overall, the two representations provide complementary information: behavioral fingerprints summarize operation and decision distributions that recur across rounds, while sequence embeddings preserve action order in combat windows. This complementarity yields a small improvement on $\mathcal{D}_{\mathrm{PER}}$ but no consistent gain on $\mathcal{D}_{\mathrm{PRO}}$. This may reflect the smaller Professional training set and redundancy between the transferred sequence representation and existing behavioral statistics.

The largest gain comes from explicit pairwise comparison. After adding explicit pairwise comparison features, E5 improves over E4 on $\mathcal{D}_{\mathrm{PER}}$ by 0.066 AUC on average (+0.049/+0.096/+0.079/+0.045/+0.036/+0.093); on $\mathcal{D}_{\mathrm{PRO}}$, the gain is 0.082 AUC (+0.089/+0.077/+0.068/+0.078/+0.055/+0.123). With explicit comparison features, the model directly uses absolute and relative differences on the same behavioral dimensions, reducing the need to learn symmetric difference relations from limited samples.

\subsubsection{Model Comparison}
\textbf{Sequence feature comparison.} The \(x^{\mathrm{seq}}\) in Eq.~\eqref{eq:sequence-embedding} in Section 3.3 is produced by a separately trained Transformer sequence encoder. The reported configuration uses continuous numerical combat-window sequences focused on shooting interactions. This encoder is trained with player identity labels, models action order around combat windows, and aggregates multiple windows into a demo-player-level sequence representation, which then enters the final pairwise input together with the game-understanding-based \(x^{\mathrm{beh}}\).

\textbf{Pairwise model comparison.} Table~\ref{tab:4} compares final-layer pairwise models on frozen outer-test predictions from $\mathcal{T}_{\mathrm{PER},pair,seq}^{(1:6)}$ and $\mathcal{T}_{\mathrm{PRO},pair,seq}^{(1:6)}$; no tuning follows these comparisons. Within each dataset, all models use exactly the same frozen E5 input representation, $x^{\mathrm{beh}} + x^{\mathrm{seq}} + \operatorname{compare}(x_i,x_j) + c_{ij}$. We compare LightGBM \cite{lightgbm-nips-2017}, XGBoost \cite{xgboost-kdd-2016}, FastMLP, and a rank-average ensemble.

\begin{table}[t]
\centering
\scriptsize
\setlength{\tabcolsep}{3.5pt}
\renewcommand{\arraystretch}{1.08}
\caption{Six-fold mean AUC for pairwise model families on the two frozen E5 evaluation surfaces.}
\label{tab:4}
\begin{tabular*}{\columnwidth}{@{\extracolsep{\fill}}llrr@{}}
\toprule
dataset & model & AUC & $\Delta$ vs. LightGBM \\
\midrule
$\mathcal{D}_{\mathrm{PER}}$ & LightGBM & \textbf{0.926} & -- \\
$\mathcal{D}_{\mathrm{PER}}$ & XGBoost & 0.918 & -0.008 \\
$\mathcal{D}_{\mathrm{PER}}$ & FastMLP & 0.858 & -0.067 \\
$\mathcal{D}_{\mathrm{PER}}$ & rank-average & 0.918 & -0.008 \\
$\mathcal{D}_{\mathrm{PRO}}$ & LightGBM & \textbf{0.956} & -- \\
$\mathcal{D}_{\mathrm{PRO}}$ & XGBoost & 0.950 & -0.005 \\
$\mathcal{D}_{\mathrm{PRO}}$ & FastMLP & 0.883 & -0.073 \\
$\mathcal{D}_{\mathrm{PRO}}$ & rank-average & 0.946 & -0.010 \\
\bottomrule
\end{tabular*}
\end{table}

LightGBM is used as the final pairwise scorer and obtains the highest six-fold mean AUC on both datasets. XGBoost and rank averaging remain close, while FastMLP is consistently weaker.

Our final implementation therefore uses game-understanding-based behavioral fingerprints, Transformer-derived sequence representations, explicit comparison features, and a LightGBM pairwise scorer.

\textbf{Implementation details.} The sequence encoder is trained for 12 epochs with AdamW (learning rate $2\times10^{-4}$, weight decay 0.02) and gradient-norm clipping at 1.0. The LightGBM scorer uses 63 leaves, learning rate 0.04, feature fraction 0.78, bagging fraction 0.85, and minimum child size 160; it trains for at most 450 rounds with validation-side early stopping after 40 rounds. ChatGPT/Codex assisted code drafting, figure preparation, and language editing; all outputs were reviewed and verified by the author. A de-identified research code package is planned for public release.

\subsection{Feature-Family Sensitivity: Which Behaviors Carry Identity Signal}
Table~\ref{tab:5} reports feature-family sensitivity on $\mathcal{T}_{\mathrm{PER},pair}^{(1:6)}$ and $\mathcal{T}_{\mathrm{PRO},pair}^{(1:6)}$, together with feature-layer analysis by behavioral mechanism on $\mathcal{T}_{\mathrm{PER},pair}^{(1)}$ and $\mathcal{T}_{\mathrm{PRO},pair}^{(1:6)}$, summarizing the features that most support the behavioral findings. Every setting in Table~\ref{tab:5} fixes the same explicit comparison input and changes only the behavioral feature families retained in $x^{\mathrm{beh}}$.

\begin{table}[t]
\centering
\tiny
\setlength{\tabcolsep}{1.4pt}
\renewcommand{\arraystretch}{1.03}
\caption{Behavioral feature sensitivity (explicit comparison fixed; sequence excluded).}
\label{tab:5}
\begin{tabular*}{\columnwidth}{@{\extracolsep{\fill}}cccc@{}}
\toprule
data & setting & eval. & AUC ($\Delta$) \\
\midrule
$\mathcal{D}_{\mathrm{PER}}$ & full $x^{\mathrm{beh}}$ & $\mathcal{T}_{\mathrm{PER},pair}^{(1:6)}$ & 0.919 (0) \\
$\mathcal{D}_{\mathrm{PER}}$ & remove aiming/crosshair & $\mathcal{T}_{\mathrm{PER},pair}^{(1:6)}$ & 0.833 (-0.087) \\
$\mathcal{D}_{\mathrm{PER}}$ & only aiming/crosshair & $\mathcal{T}_{\mathrm{PER},pair}^{(1:6)}$ & 0.890 (-0.029) \\
$\mathcal{D}_{\mathrm{PER}}$ & remove mechanics/state & $\mathcal{T}_{\mathrm{PER},pair}^{(1:6)}$ & 0.908 (-0.012) \\
$\mathcal{D}_{\mathrm{PRO}}$ & full $x^{\mathrm{beh}}$ & $\mathcal{T}_{\mathrm{PRO},pair}^{(1:6)}$ & 0.955 (0) \\
$\mathcal{D}_{\mathrm{PRO}}$ & remove aiming/crosshair & $\mathcal{T}_{\mathrm{PRO},pair}^{(1:6)}$ & 0.880 (-0.076) \\
$\mathcal{D}_{\mathrm{PRO}}$ & only aiming/crosshair & $\mathcal{T}_{\mathrm{PRO},pair}^{(1:6)}$ & 0.915 (-0.041) \\
$\mathcal{D}_{\mathrm{PRO}}$ & remove mechanics/state & $\mathcal{T}_{\mathrm{PRO},pair}^{(1:6)}$ & 0.950 (-0.006) \\
\midrule
$\mathcal{D}_{\mathrm{PER}}$ & full $x^{\mathrm{beh}}$ & $\mathcal{T}_{\mathrm{PER},pair}^{(1)}$ & 0.916 (0) \\
$\mathcal{D}_{\mathrm{PER}}$ & only low-level operations & $\mathcal{T}_{\mathrm{PER},pair}^{(1)}$ & 0.915 (-0.001) \\
$\mathcal{D}_{\mathrm{PER}}$ & remove low-level operations & $\mathcal{T}_{\mathrm{PER},pair}^{(1)}$ & 0.738 (-0.179) \\
$\mathcal{D}_{\mathrm{PER}}$ & remove rhythm/space & $\mathcal{T}_{\mathrm{PER},pair}^{(1)}$ & 0.914 (-0.002) \\
$\mathcal{D}_{\mathrm{PER}}$ & remove tactics/context & $\mathcal{T}_{\mathrm{PER},pair}^{(1)}$ & 0.914 (-0.002) \\
$\mathcal{D}_{\mathrm{PRO}}$ & full $x^{\mathrm{beh}}$ & $\mathcal{T}_{\mathrm{PRO},pair}^{(1:6)}$ & 0.955 (0) \\
$\mathcal{D}_{\mathrm{PRO}}$ & only low-level operations & $\mathcal{T}_{\mathrm{PRO},pair}^{(1:6)}$ & 0.947 (-0.009) \\
$\mathcal{D}_{\mathrm{PRO}}$ & remove low-level operations & $\mathcal{T}_{\mathrm{PRO},pair}^{(1:6)}$ & 0.842 (-0.113) \\
$\mathcal{D}_{\mathrm{PRO}}$ & remove rhythm/space & $\mathcal{T}_{\mathrm{PRO},pair}^{(1:6)}$ & 0.953 (-0.003) \\
$\mathcal{D}_{\mathrm{PRO}}$ & remove tactics/context & $\mathcal{T}_{\mathrm{PRO},pair}^{(1:6)}$ & 0.952 (-0.003) \\
\bottomrule
\end{tabular*}
\end{table}

Table~\ref{tab:5} shows that aiming/crosshair is the strongest identity signal. On $\mathcal{D}_{\mathrm{PER}}$, the six-fold average AUC of the full $x^{\mathrm{beh}}$ is 0.919, and removing aiming/crosshair lowers it by 0.087 on average (-0.114/-0.090/-0.098/-0.053/-0.068/-0.096); on $\mathcal{D}_{\mathrm{PRO}}$, the corresponding AUC is 0.955 and the average decrease is 0.076 (-0.082/-0.068/-0.046/-0.063/-0.062/-0.134). Using only aiming/crosshair reaches AUCs of 0.890 and 0.915, respectively; among the reported removals, mechanics/state provides the next-largest contribution.

Merging the eight feature types by behavioral mechanism concentrates the conclusion: on $\mathcal{T}_{\mathrm{PER},pair}^{(1)}$ and $\mathcal{T}_{\mathrm{PRO},pair}^{(1:6)}$, using only aiming/crosshair, mechanics/state, and combat/engagement lowers AUC by only 0.001 and 0.009, respectively, while removing these three lowers AUC by 0.179 and 0.113. The two datasets yield the same conclusion: CS2 player identity signals come mainly from crosshair control, combat micro-operations, and movement-stop-fire coordination, while rhythm, space, buying, and context preferences provide supplementary information.

\subsection{Account-History Aggregation: From Pairwise Scores to Multi-Demo History Comparison}
Account-history review compares a current demo against multiple historical demos. Table~\ref{tab:6} reports account-history aggregation on $\mathcal{T}_{\mathrm{PER},hist}^{(1:6)}(K)$ and $\mathcal{T}_{\mathrm{PRO},hist}^{(1:6)}(K)$. Each eligible query forms one positive group from $K$ strictly earlier observations of the same identity/account key and four negative groups, each formed from one different identity candidate; deterministic, score-blind nested prefixes are fixed by each dataset contract. The mean LLR in Eq.~\eqref{eq:llr-mean} aggregates each group's $K$ scores.

\begin{table}[!tbp]
\centering
\tiny
\setlength{\tabcolsep}{1.2pt}
\renewcommand{\arraystretch}{1.02}
\caption{Account-history aggregation as history depth $K$ changes.}
\label{tab:6}
\begin{tabular*}{\columnwidth}{@{\extracolsep{\fill}}cccl@{}}
\toprule
data & $K$ & mean-LLR AUC & six-fold AUCs \\
\midrule
$\mathcal{D}_{\mathrm{PER}}$ & 1 & 0.923 (baseline) & 0.935/0.918/0.878/0.940/0.949/0.918 \\
$\mathcal{D}_{\mathrm{PER}}$ & 3 & 0.966 (+0.043) & 0.976/0.960/0.938/0.962/0.985/0.974 \\
$\mathcal{D}_{\mathrm{PER}}$ & 5 & 0.975 (+0.052) & 0.981/0.967/0.955/0.975/0.989/0.983 \\
$\mathcal{D}_{\mathrm{PER}}$ & 10 & 0.982 (+0.058) & 0.986/0.975/0.966/0.982/0.992/0.987 \\
$\mathcal{D}_{\mathrm{PRO}}$ & 1 & 0.914 & 0.893/0.941/0.922/0.919/0.906/0.905 \\
$\mathcal{D}_{\mathrm{PRO}}$ & 3 & 0.966 & 0.970/0.969/0.962/0.967/0.968/0.959 \\
$\mathcal{D}_{\mathrm{PRO}}$ & 5 & 0.975 & 0.974/0.983/0.969/0.980/0.978/0.968 \\
$\mathcal{D}_{\mathrm{PRO}}$ & 10 & not estimable & no eligible $K=10$ queries in any fold \\
\bottomrule
\end{tabular*}
\end{table}

Multi-demo history comparison provides a more stable account-level signal than a single pairwise score. To keep $K$ comparisons paired, all Perfect rows use the same fixed $K=10$-eligible cohort (3,782 query-fold instances), while Professional $K=1,3,5$ use the same fixed $K=5$-eligible cohort (446 query-fold instances). On $\mathcal{D}_{\mathrm{PER}}$, mean-LLR AUC rises from 0.923 at $K=1$ to 0.966 at $K=3$ and 0.982 at $K=10$; on $\mathcal{D}_{\mathrm{PRO}}$, the corresponding $K=1,3,5$ values are 0.914, 0.966, and 0.975. No Professional fold has an eligible $K=10$ query. These fixed history-group cohorts differ from the pair surface in Table~\ref{tab:3}; on Perfect, raw-score mean gives very similar AUC, indicating that the main gain comes from accumulating multiple evidence items.

\subsection{Cross-Time, Cross-Map, and Data Expansion Experiments}
\subsubsection{Time-Gap Sensitivity}
Actual review often occurs with larger time gaps: a platform or tournament organizer obtains a recent suspicious demo and compares it with earlier historical demos of the account. This setting is harder than random pairs, because player state, map pool, version, settings, and play style may all change over time.

$\mathcal{D}_{\mathrm{PER}}$ observations span April 14 to August 19, 2026, forming same-day, 1--7-day, 8--30-day, 31--90-day, and over-90-day comparisons; the available $\mathcal{D}_{\mathrm{PRO}}$ comparisons cover same-day, 1--7-day, and 8--30-day gaps.

\begin{table}[!tbp]
\centering
\scriptsize
\setlength{\tabcolsep}{2pt}
\renewcommand{\arraystretch}{1.05}
\caption{Time-gap sensitivity on the two datasets.}
\label{tab:7}
\begin{tabular*}{\columnwidth}{@{\extracolsep{\fill}}cccc@{}}
\toprule
data & time gap & pairs & AUC \\
\midrule
$\mathcal{D}_{\mathrm{PER}}$ & same day & 187,171 & 0.985 \\
$\mathcal{D}_{\mathrm{PER}}$ & 1--7 days & 321,023 & 0.944 \\
$\mathcal{D}_{\mathrm{PER}}$ & 8--30 days & 345,533 & 0.927 \\
$\mathcal{D}_{\mathrm{PER}}$ & 31--90 days & 220,265 & 0.885 \\
$\mathcal{D}_{\mathrm{PER}}$ & $>90$ days & 444,338 & 0.894 \\
$\mathcal{D}_{\mathrm{PRO}}$ & same day & 29,626 & 0.989 \\
$\mathcal{D}_{\mathrm{PRO}}$ & 1--7 days & 45,126 & 0.949 \\
$\mathcal{D}_{\mathrm{PRO}}$ & 8--30 days & 32,938 & 0.924 \\
$\mathcal{D}_{\mathrm{PRO}}$ & $\geq31$ days & 0 & -- \\
\bottomrule
\end{tabular*}
\end{table}

Table~\ref{tab:7} shows that longer time gaps are generally harder: AUC on $\mathcal{D}_{\mathrm{PER}}$ declines from 0.985 for same-day comparisons to 0.885 for 31--90 days and is 0.894 beyond 90 days; on $\mathcal{D}_{\mathrm{PRO}}$, it declines from 0.989 for same-day comparisons to 0.924 for 8--30 days. This is consistent with intuition: over longer gaps, the same player's state, map pool, and play style may change, yet the model retains useful cross-time recognition.

\subsubsection{Cross-Map Robustness}
Cross-map comparisons are common in account-history review and remove some map-specific contextual similarity.

\begin{table}[!tbp]
\centering
\tiny
\setlength{\tabcolsep}{1pt}
\renewcommand{\arraystretch}{1.04}
\caption{Same-map vs cross-map sensitivity on the two datasets.}
\label{tab:8}
\begin{tabular}{@{}clrrcc@{}}
\toprule
data & relation & pairs & same/diff. & AUC/$\Delta$ & same-AP/$\Delta$ \\
\midrule
$\mathcal{D}_{\mathrm{PER}}$ & all & 1,518,330 & 153,790/1,364,540 & 0.926/-- & 0.703/-- \\
$\mathcal{D}_{\mathrm{PER}}$ & same & 462,043 & 41,862/420,181 & 0.951/+0.026 & 0.769/+0.066 \\
$\mathcal{D}_{\mathrm{PER}}$ & cross & 1,056,287 & 111,928/944,359 & 0.914/-0.012 & 0.676/-0.027 \\
$\mathcal{D}_{\mathrm{PRO}}$ & all & 107,690 & 10,769/96,921 & 0.956/-- & 0.775/-- \\
$\mathcal{D}_{\mathrm{PRO}}$ & same & 38,295 & 1,823/36,472 & 0.980/+0.024 & 0.808/+0.034 \\
$\mathcal{D}_{\mathrm{PRO}}$ & cross & 69,395 & 8,946/60,449 & 0.942/-0.014 & 0.768/-0.007 \\
\bottomrule
\end{tabular}
\end{table}

Table~\ref{tab:8} shows that on $\mathcal{D}_{\mathrm{PER}}$, same-map AUC is 0.026 above the all-pair result and cross-map AUC is 0.012 below it; on $\mathcal{D}_{\mathrm{PRO}}$, the differences are +0.024 and -0.014. Because sampled same-demo negatives are necessarily same-map, part of the same-map advantage reflects pair construction. Cross-map AUCs remain 0.914 and 0.942.

\subsubsection{Cross-Dataset Training Between Perfect and Professional}
We examine the effect of training-data source on $\mathcal{T}_{\mathrm{PER},pair}^{(1:6)}$ and $\mathcal{T}_{\mathrm{PRO},pair}^{(1:6)}$.

\begin{table}[t]
\centering
\scriptsize
\setlength{\tabcolsep}{3pt}
\caption{Cross-dataset training under a separate frozen augmentation protocol ($\Delta$ relative to $\mathcal{D}_{\mathrm{PER}}$-only training within each test block).}
\label{tab:9}
\begin{tabular}{@{}llrr@{}}
\toprule
test & training data & AUC & $\Delta$ vs. block baseline \\
\midrule
$\mathcal{D}_{\mathrm{PER}}$ & $\mathcal{D}_{\mathrm{PER}}$ & 0.911 & baseline \\
$\mathcal{D}_{\mathrm{PER}}$ & $\mathcal{D}_{\mathrm{PER}}+\mathcal{D}_{\mathrm{PRO}}$ & 0.914 & +0.003 \\
\addlinespace[2pt]
$\mathcal{D}_{\mathrm{PRO}}$ & $\mathcal{D}_{\mathrm{PER}}$ & 0.912 & baseline \\
$\mathcal{D}_{\mathrm{PRO}}$ & $\mathcal{D}_{\mathrm{PRO}}$ & \textbf{0.955} & +0.043 \\
$\mathcal{D}_{\mathrm{PRO}}$ & $\mathcal{D}_{\mathrm{PER}}+\mathcal{D}_{\mathrm{PRO}}$ & 0.948 & +0.036 \\
\bottomrule
\end{tabular}
\end{table}

Table~\ref{tab:9} uses a separate frozen cross-dataset augmentation protocol; its $\mathcal{D}_{\mathrm{PER}}$-only baseline uses a different training-pair construction from the within-dataset ablation baseline in Table~\ref{tab:5}. On $\mathcal{T}_{\mathrm{PRO},pair}^{(1:6)}$, Professional-only training reaches AUC 0.955, 0.043 above the 0.912 from Perfect-only training; mixed training reaches 0.948. On the Perfect test surface, adding Professional training data changes AUC from 0.911 to 0.914. The lower mixed-versus-Professional-only result may reflect differences between the two data domains in team roles, match intensity, and behavior distributions.

\subsubsection{External Test on 5E}
$\mathcal{D}_{5\mathrm{E}}$ uses exact SteamID as a weak same-account label, so the result cannot be directly interpreted as same-natural-person verification. The behavior-and-explicit-comparison model trained on $\mathcal{D}_{\mathrm{PER}}$ reaches zero-shot AUC 0.966 on the full fixed test surface $\mathcal{T}_{5\mathrm{E}}$ (55,840 pairs), providing evidence of cross-platform transfer under weak same-account labels. After excluding one SteamID that overlaps the training source, $\mathcal{T}_{5\mathrm{E}}^{\mathrm{disjoint}}$ retains 55,676 pairs (99.7\%) and the AUC remains 0.966.

\FloatBarrier
\section{Discussion and Limitations}
\subsection{Evaluation Boundaries}
Our primary splits are formed by the constructed identity ledger and assign each demo to one side. Under this ledger, known natural-person, SteamID, alias, observation, demo, and content overlaps between training and test are zero. The model input is restricted to per-player fingerprints and a same-map/cross-map flag; it excludes demo IDs, match IDs, teammate/opponent identities, and shared-match identifiers. Undisclosed cross-account ownership may still violate true person disjointness and is treated as residual label noise.

Using 1,000 player-cluster bootstrap replicates, the final model has 95\% AUC confidence intervals of [0.915, 0.950] on $\mathcal{D}_{\mathrm{PER}}$ and [0.948, 0.965] on $\mathcal{D}_{\mathrm{PRO}}$, indicating that the results are not driven by a few high-contribution players.

After excluding different-player pairs drawn from the same demo, E5 AUC remains 0.920 on Perfect and 0.944 on Professional, indicating that performance is not driven by same-match negatives.

\subsection{Data and Label Boundaries}
In $\mathcal{D}_{\mathrm{PER}}$, same-player labels come from players' manual confirmation of account histories, account sharing, and multi-account ownership; undisclosed borrowing, temporary substitution, or account sharing may still introduce label noise. The results therefore depend on the completeness of the manual confirmations, and deployment should retain identity audits and feedback from new evidence.

\subsection{Practical Deployment: Runtime Cost and Responsible Use}

Once fingerprints and embeddings are available, pairwise scoring and history aggregation process about 24k pairs/s on Apple M4; demo parsing and feature extraction remain the dominant offline cost.

In deployment, a new demo can be compared with historical observations to prioritize cases where current behavior is clearly inconsistent with the account history.

On $\mathcal{D}_{\mathrm{PER}}$, retrieved account histories may include observations from another operator. On the $K=5$ subset for which eligible third-player replacements can be constructed, 3,674/3,782 queries (97.1\%) are retained. Replacing one, two, or three of the five positive histories with different-player observations lowers mean-LLR AUC from 0.975 to 0.963, 0.941, and 0.894, respectively.

\section{Conclusion and Future Work}
We formulate account-history consistency review on competitive FPS platforms as open-set same-player verification, using CS2 demos to design per-demo-player behavioral fingerprints and learn pairwise consistency.

Experiments show strong discrimination: on the sequence-common Perfect and Professional evaluation surfaces, the full model reaches ROC AUCs of 0.926 and 0.956. Identity signal comes mainly from aiming/crosshair and other low-level mechanical behaviors. On fixed eligible query cohorts, mean-LLR account-history AUC rises from 0.923 at $K=1$ to 0.982 at $K=10$ on Perfect and from 0.914 at $K=1$ to 0.975 at $K=5$ on Professional.

Future work will study longer-term drift, partial-match verification, and human-in-the-loop review.

\section{Ethical Considerations}
This work is intended to provide identity-consistency evidence for prioritizing account-history review, rather than to determine player identity or impose automated sanctions. False positives may subject legitimate players to unwarranted suspicion, while sparse histories, hardware or setting changes, and atypical play styles may affect model scores. Any operational use should therefore combine multiple sources of evidence with human review, appeals, threshold calibration, and continuing audits, and should not treat a single model score as grounds for enforcement.

The gameplay demos analyzed in this study were publicly accessible. For the manually confirmed Perfect subset, participating players were informed that their demos and identity confirmations would be used for model training and research, and they consented to this research use and to the release of de-identified derived features.

Game demos contain fine-grained behavioral trajectories, and learned fingerprints could be repurposed for unwanted tracking or profiling. The model inputs exclude real names, SteamIDs, demo IDs, match IDs, and other direct identifiers; we report aggregate results, and the de-identified research artifact excludes raw demos, identity mappings, and identity ledgers. Storage, access, and subsequent sharing of manually confirmed information, public professional-match records, and derived representations should follow data-minimization principles and the scope of the original authorization.

The evaluated data cover particular platforms, player communities, and professional matches, and do not establish equal performance across regions, skill levels, hardware environments, or long-term behavioral drift. Deliberate imitation or behavior modification may also evade review, and this work does not establish robustness under real adversarial conditions. Deployment should monitor error rates across populations and use cases and constrain the system's purpose accordingly.

\end{document}